\documentclass{article} % For LaTeX2e
\usepackage{iclr2027_conference,times}

\usepackage{amsmath,amsfonts,bm}

\def\eqref#1{equation~\ref{#1}}
\def\1{\bm{1}}

\DeclareMathAlphabet{\mathsfit}{\encodingdefault}{\sfdefault}{m}{sl}
\SetMathAlphabet{\mathsfit}{bold}{\encodingdefault}{\sfdefault}{bx}{n}

\usepackage[
  colorlinks=true,
  linkcolor=red!25!blue,   % internal links (sections, equations, theorems)
  citecolor=red!25!blue,   % bibliography citations
  urlcolor=red!25!blue,    % \url and \href links
  filecolor=red!25!blue    % links to local files
]{hyperref}
\usepackage{url}
\usepackage{tikz}
\usetikzlibrary{calc}
\usetikzlibrary{patterns.meta}
\usepackage{makecell}
\usepackage{amssymb}
\usepackage[table]{xcolor}
\usepackage{multirow}
\usepackage{xspace}
\usepackage{caption}
\usepackage{etoolbox}
\AtBeginEnvironment{table}{\renewenvironment{center}{\centering}{\par}}

\usepackage{wrapfig}
\usepackage{microtype}
\usepackage{enumitem}

\newcommand{\ctrlmcp}{\textsc{control}\xspace}
\newcommand{\rocqmcp}{\textsc{rocq-mcp}\xspace}
\newcommand{\rme}{\textsc{rocq-mcp-evolve}\xspace}
\newcommand{\lme}{\textsc{lean-mcp-evolve}\xspace}
\newcommand{\leanmcp}{\textsc{lean-lsp-mcp}\xspace}

\newcommand{\orch}{orchestrator\xspace}
\newcommand{\testers}{testers\xspace}
\newcommand{\Testers}{Testers\xspace}

\newcommand{\tool}[1]{\texttt{#1}}

\newcommand{\buck}[3]{{\color{gray}\small(#1\,/\,#2\,/\,#3)}}
\newcommand{\midsep}{\hspace{0.4em}$\mid$\hspace{0.4em}}
\newlength{\grpsepwidth}
\newcommand{\grpsep}{\hspace{\grpsepwidth}}

\definecolor{revertedcolor}{RGB}{235,235,235}
\newcommand{\keptrow}{}
\newcommand{\revertedrow}{\rowcolor{revertedcolor}}
\newcommand{\gr}[1]{{\leavevmode\color{gray}#1}}  % gray cell text, for reverted rows
\makeatletter
\newdimen\bugtabwidth
\newcommand{\bugrow}{\tikz[overlay]\path[pattern={Lines[angle=45, distance=3pt, line width=0.4pt]}, pattern color=gray!60]
  (-\tabcolsep,-\dp\@arstrutbox) rectangle (\bugtabwidth-\tabcolsep,\ht\@arstrutbox);}
\newcommand{\bugtabular}[1]{%
  \def\bugtabbody{#1}%
  \setbox0\hbox{\let\bugrow\relax\bugtabbody}%
  \global\bugtabwidth=\wd0
  \bugtabbody}
\makeatother

\title{Growing an Agent/Prover Interface: Evolutionary Tool Design for Cost-Efficient Theorem Proving in Rocq and Lean}

\author{
Jules Viennot \\
IRIF, Université Paris Cité, Inria, CNRS
\And
Guillaume Baudart \\
IRIF, Université Paris Cité, Inria, CNRS
\And
Marc Lelarge \\
DI ENS, PSL University, Inria
}

\iclrfinalcopy % Uncomment for camera-ready version, but NOT for submission.

\begin{document}

\maketitle
\lhead{}
\renewcommand{\headrulewidth}{0pt}

\begin{abstract}
Recent achievements in AI-assisted mathematics require intensive interaction of agents with proof assistants to generate machine-checked proof certificates.
Agents interact with proof assistants such as Rocq or Lean through an interface that controls what the agent receives from the prover and the cost of these interactions.
Today, these interfaces are adapted from tools designed for humans and not optimized for agents.
We propose an evolutionary method where a frontier model incrementally proposes new features and only keeps the ones that improve the overall performance of smaller models.
We demonstrate the effectiveness of our method by growing, on a curated set of mathematical problems, \rme, a new MCP server for the Rocq prover.
On the held-out \texttt{test} split of miniF2F-Rocq, an agent equipped with \rme outperforms both the baseline that only exposes the Rocq compiler and an established MCP server, across four models from two families, in success rate, cost per solve, and time per solve.
Although evolved for Rocq, the resulting server transfers to Lean, improving cost and time per solve on a subset of PutnamBench.
We release \rme and its port to Lean.
\end{abstract}

\begin{figure}[h]
\begin{center}
% Final results: 3 evenly-spaced bar diagrams sharing one legend above them.
\input{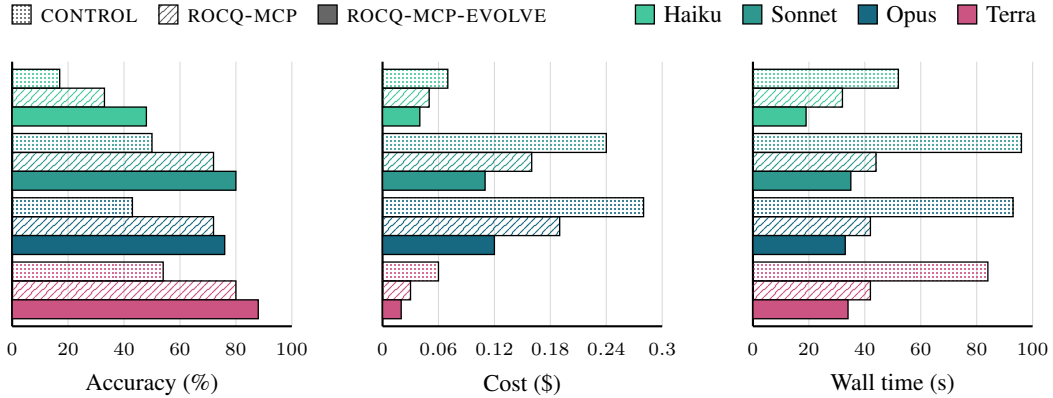}

\definecolor{haikucolor}{HTML}{44C79C}
\definecolor{sonnetcolor}{HTML}{2E988F}
\definecolor{opuscolor}{HTML}{176982}
\definecolor{terracolor}{HTML}{CB5482}

\begin{tikzpicture}

  % --- Models ---
  \def\models{Haiku/haikucolor,Sonnet/sonnetcolor,Opus/opuscolor,Terra/terracolor}
  \listlength{\models}{\nmodels}

  % --- Toolings ---
  \def\toolings{\ctrlmcp /dotted, \rocqmcp /striped, \rme /full}
  \listlength{\toolings}{\ntoolings}

  % --- Data ---
  % Pooled (whole-dataset) accuracy/cost/wall time, per model, in
  % \toolings order (\ctrlmcp/\rocqmcp/\rme), from the "pooled" column of
  % Table~\ref{tab:detailed_results} (figures/detailed_results.tex).
  \def\dataA{{17, 33, 48},{50, 72, 80},{43, 72, 76},{54, 80, 88}}
  \def\dataB{{0.07, 0.05, 0.04},{0.24, 0.16, 0.11},{0.28, 0.19, 0.12},{0.06, 0.03, 0.02}}
  \def\dataC{{52, 32, 19},{96, 44, 35},{93, 42, 33},{84, 42, 34}}

  % --- Diagrams geometry ---
  % Bars are horizontal: models stack downward, values grow rightwards over
  % \barchartlength; the diagrams sit side by side.
  \def\barchartlength{3.7}
  \def\barwidth{0.25}     % bar thickness
  \def\bargroupgap{.1}
  \def\diagramgap{1.2}
  \barchartnbars{\dataA}{\nbars}
  \pgfmathsetmacro{\diagramspan}{\nbars*\barwidth+(\nmodels+1)*\bargroupgap}
  \pgfmathsetmacro{\diagramoffsetB}{\barchartlength+\diagramgap}
  \pgfmathsetmacro{\diagramoffsetC}{2*(\barchartlength+\diagramgap)}

  % --- Legend ---
  % Layout knobs, all in cm and all local to this picture (\renewcommand
  % inside the tikzpicture's group doesn't leak into other figures):
  %   \legendx, \legendy : where the row starts (bottom-left of the first swatch)
  %   \barlegendentrygap : space after a label, before the next swatch
  %   \barlegendgroupgap : extra space where \barlegendgroupbreak sits, i.e.
  %                        between the models group and the toolings group
  % (Values below equal the library defaults, so the layout is unchanged.)
  \def\legendx{0}
  \def\legendy{0.5}
  \renewcommand{\barlegendentrygap}{0.15}
  \renewcommand{\barlegendgroupgap}{0.9}
  \barlegendreset[\legendx]{\legendy}
  \foreach \tool/\pat in \toolings {
    \barlegendpatternentry{\pat}{\tool}
  }
  \barlegendgroupbreak
  \foreach \model/\col in \models {
    \barlegendentry{\col}{\model}
  }

  % --- Diagram 1: Benchmark A ---
  % vmin lowered from 40: Haiku's \ctrlmcp accuracy (16) is well below that.
  \def\vmin{0}
  \def\vmax{100}
  \begin{scope}
    \bargridlinesN{\diagramspan}{\vmin}{\vmax}{5}
    \baraxes{\diagramspan}
    \baraxistitle{\diagramspan}{Accuracy (\%)}
    \bargroupedbars{\models}{\dataA}{\toolings}{\vmin}{\vmax}{\barwidth}{\bargroupgap}
  \end{scope}

  % --- Diagram 2: Benchmark B ---
  \def\vmin{0}
  \def\vmax{0.3}
  \begin{scope}[xshift=\diagramoffsetB cm]
    \bargridlinesN{\diagramspan}{\vmin}{\vmax}{5}
    \baraxes{\diagramspan}
    \baraxistitle{\diagramspan}{Cost (\$)}
    \bargroupedbars{\models}{\dataB}{\toolings}{\vmin}{\vmax}{\barwidth}{\bargroupgap}
  \end{scope}

  % --- Diagram 3: Benchmark C ---
  % vmin lowered from 25: Haiku's \rme wall time (17) is below that.
  \def\vmin{0}
  \def\vmax{100}
  \begin{scope}[xshift=\diagramoffsetC cm]
    \bargridlinesN{\diagramspan}{\vmin}{\vmax}{5}
    \baraxes{\diagramspan}
    \baraxistitle{\diagramspan}{Wall time (s)}
    \bargroupedbars{\models}{\dataC}{\toolings}{\vmin}{\vmax}{\barwidth}{\bargroupgap}
  \end{scope}
\end{tikzpicture}
\end{center}
\caption{
  Results on the \texttt{test} split of miniF2F-Rocq.
  Each model is evaluated with three MCP servers: \ctrlmcp, a minimal server wrapping the Rocq compiler, \rocqmcp, an established MCP server for Rocq, and \rme.
  For a fair comparison, cost and wall time are averaged over the theorems proved with all three servers.
}
\label{fig:intro_results}
\end{figure}

\pagebreak

\section{Introduction}
\label{sec:intro}

% Domain presentation
Large Language Models (LLMs) have made rapid progress in mathematical reasoning with impressive public results~\citep{navier-stokes, riemann-zeta, unit-distance}.
However, trust and audit of the generated proofs remain a major concern.
Proof assistants such as Lean~\citep{lean}, Isabelle~\citep{isabelle}, and Rocq~\citep{rocq} emerged as natural validation tools to mitigate this issue.
Major announcements are now backed by machine-checked proof certificates~\citep{fermat, navier-stokes, riemann-zeta}.
The generation of these certificates requires intensive interaction between the agents and the proof assistant.

Agents interact with the proof assistant through a specialized interface that controls what the agent receives from the prover.
There exist several specialized interfaces for proof assistants~\citep{rocq-mcp,lean-lsp-mcp,pantograph,axle,numina-lean-agent}, most implementing the new Model Context Protocol (MCP) standard.
An MCP server exposes a series of tools for the agents.
For proof assistants, tools range from compiling proofs to interactive debugging and searching for relevant theorems.
Today, these interfaces are adapted from systems designed for humans and not optimized for agents.
Yet, these interfaces are critical: they drive the cost of each interaction and new generations of models are trained against them~\citep{leanstral}.

We propose an evolutionary method where a frontier model incrementally proposes new features and only keeps the ones that improve the overall performance of smaller models.
We used Claude Fable~5 to drive the experiment and Claude Sonnet~5 and Claude Haiku~4.5 to evaluate each feature on a fixed set of problems.
Starting from a minimal interface, the frontier model proposes new features one by one.
At each step, we evaluate the performance gap between two consecutive versions and only keep features that improve our global objectives: (1)~\textit{accuracy}: the proportion of problems solved, (2)~\textit{cost}: the cost per problem solved, and (3)~\textit{wall time}: the wall-clock time per problem solved.

We demonstrate the effectiveness of our method by growing \rme, a new MCP server for the Rocq prover, starting from a server that only exposes the Rocq compiler.
At each step of the evolution, we evaluate agents on mathematical problems from the valid split of miniF2F-Rocq~\citep{minif2frocq} and the Rocq Workbook~\citep{rocq-workbook}, and five project-scale tasks.
We focus on the Rocq prover to mitigate data contamination of the main benchmarks: miniF2F~\citep{minif2f} and PutnamBench~\citep{putnambench}.
In addition, since Lean is now a popular tool in the ML community, we cannot rule out that frontier models have been trained to interact with existing interfaces.

We evaluate \rme on the disjoint \texttt{test} split of miniF2F-Rocq and compare three MCP servers across four models: (1) \ctrlmcp, which exposes only the Rocq compiler, (2) \rocqmcp, an established MCP server for Rocq~\citep{putnam2025,certigc-2026,agentic-tt-2026}, and (3) \rme.
The four models used are Claude Haiku~4.5, Claude Sonnet~5, Claude Opus~4.8, and GPT-5.6 Terra.
Figure~\ref{fig:intro_results} shows that \rme outperforms the other MCP servers for our three metrics.
Past competition exercises, we also show that \rme outperforms the other MCP servers on real-world case studies where agents are tasked with building entire projects with multiple files on top of existing libraries.
Finally, we port the resulting set of tools to Lean, and show that while the overall solve rate is lower than with the state-of-the-art \leanmcp~\citep{lean-lsp-mcp}, our server improves both cost and wall time per solve.
Both servers and all the code and experiments are available here\footnote{\url{https://github.com/LLM4Rocq/rocq-mcp-evolve-experiment}}.

\textbf{Contributions} We present the following contributions:
\begin{itemize}[itemsep=0pt]
  \item An evolutionary method to grow an MCP server for interactive theorem proving.
  \item \rme, the MCP server produced by this method, which outperforms an established interface for Rocq in solve rate, cost, and wall time.
  \item An evaluation across different model sizes and families on multiple tasks ranging from mathematical problems to project-scale tasks.
  \item A port of the server to Lean which improves cost and wall time per solve.
\end{itemize}

\section{The evolutionary process}
\label{sec:evolution}

Agents are trained to rely on external tools to complete their tasks.
For proof assistants, these tools are inherited from existing interfaces originally designed for human users.
In interactive theorem provers, a user builds a formal proof step by step using commands called \emph{tactics} to simplify the proof goal until there is nothing left to prove.
Specialized tools were introduced to help users navigate through the proof: inspect the current proof goal, search for applicable theorems, print a definition or a notation, etc.
The set of tools directly influences users' and agents' capabilities at theorem proving.
We propose to leverage the relative cost and efficiency of each tool as an objective to optimize the design of a new interface for interactive theorem proving.

\begin{figure}[t]
\centering
\includegraphics[width=\linewidth]{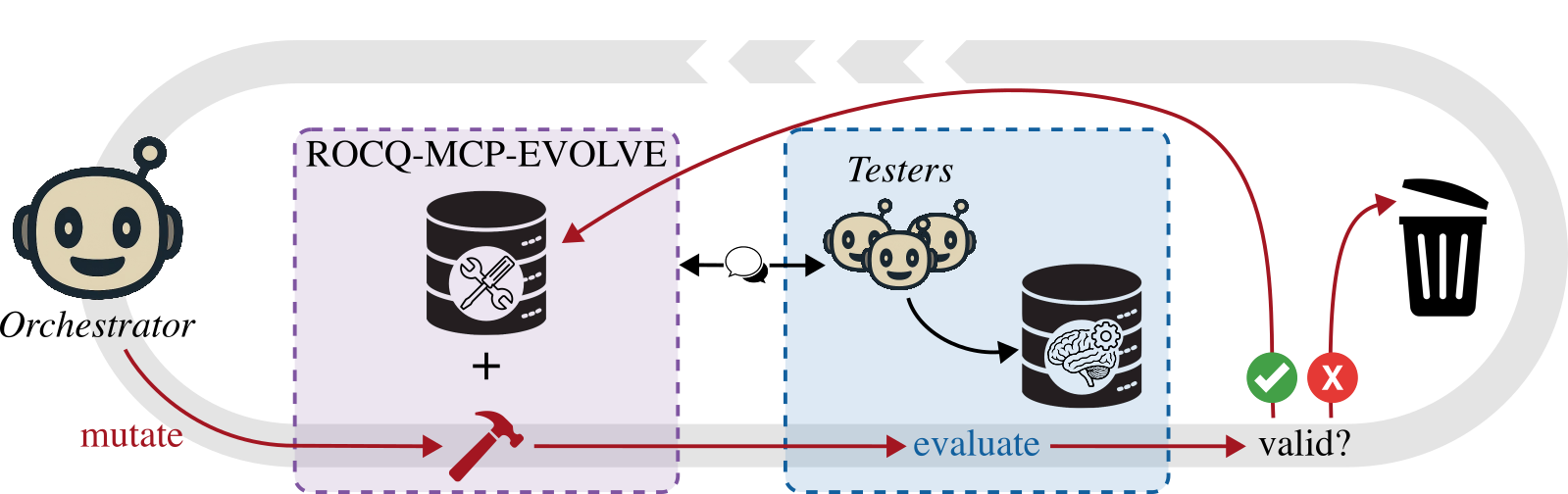}
\caption{
A step of the evolutionary process.
The mutation designed by the \orch is denoted by the dark red color.
The purple dotted box represents \rme, the blue dotted box represents the evaluation of the mutation.
The light gray circle shows the process is iterative: a new step starts at the end of the previous one.}
\label{fig:evolution_diagram}
\end{figure}

The evolutionary process starts from a \ctrlmcp MCP server that only exposes a single tool: compile an entire file.
This server is the minimal, and least interactive, interface for a proof assistant.
Then, at each step a frontier model \emph{mutates} the current MCP server by proposing a new feature: adding a new tool, refining the output of an existing one, changing the configuration, etc.
The frontier model then \emph{evaluates} each mutation with smaller models on curated datasets, and \emph{validates} or discards the mutation by measuring its added value with respect to the previous iteration.
Figure~\ref{fig:evolution_diagram} illustrates a step of the evolutionary process.

We call the frontier model that manages the entire process the \emph{\orch} and instantiate it with Claude Fable~5.
To devise and implement new mutations, the \orch leverages statistics and logs from previous runs.
During the experiment, we interact only with the \orch and our involvement is limited to setting up the environment (i.e., defining the objectives, budgets, and validation rules), and a daily review.
There is no human intervention during the design and implementation of the mutations.

We evaluate each iteration of the MCP server on curated datasets of problems.
For each problem, we launch agents equipped with the mutated version of the server, the \textit{\testers}, with a fixed budget of time and tool calls.
The score on a dataset is averaged over several runs.
\Testers are instantiated with Claude Haiku~4.5 and Claude Sonnet~5.

We track three metrics: (1)~\textit{accuracy}, the pass@1 proportion of problems solved, (2)~\textit{cost}, the average cost in US dollars per solved problem, and (3)~\textit{wall time}, the average wall-clock time per solved problem.
In the following, cost and wall time refer to averages over solved problems unless stated otherwise.
The evolutionary process comprises two phases.

% ------------------------------------------------------------------------------

\subsection{Phase~1: solving mathematical exercises}

In the first phase, we evaluate the interface with mathematical exercises.
The goal is to optimize the server for standalone theorem proving in a frozen environment.
\Testers are Haiku instances with two runs per problem, each limited to 300\,s of wall-clock time and 30 calls to the server.
They receive a system prompt that presents the toolset and provides general instructions and tips for Rocq.
An independent set of Sonnet instances is used to track regression with a larger model and avoid overspecialization of the server to Haiku.
We set a 6-day deadline to complete Phase~1.
The winning configuration was discovered on the third day.

\textbf{Datasets}
Most mutations are evaluated on dev60, a set of 60 problems from the Rocq Workbook, a translation of 10\,000 theorems from the Lean Workbook into Rocq.
Problems are categorized into three difficulty buckets: easy, medium, and hard.
The difficulty of each problem is estimated with a simple heuristic (statement length and symbol count).
We selected 20 problems per difficulty bucket.

The dev60 dataset is too limited to evaluate some mutations.
We also use three additional datasets to evaluate specific mutations: hard70 (70 hard problems from the Rocq Workbook) to test agent collaboration on difficult problems, mathcomp35 (35 lemmas extracted from the MathComp library and proved in the context of their file) to test library-specific hints and lemma retrieval, and the valid split of miniF2F-Rocq~\citep{minif2frocq} (244 problems drawn from high-school competitions) to evaluate the preloading of automation tactics.
For miniF2F, we infer the difficulty of the problems from their source: MATH problems are easy (130 problems), AMC12 and textbook exercises medium (79 problems), and AIME and IMO problems hard (35 problems).

\textbf{Validation}
The results of the \testers are compared against the results of the previous iteration on dev60 over two runs, i.e., 40 attempts per difficulty bucket.
In practice, a mutation is accepted if the net gain (new successes minus new failures) reaches $+2$ in at least one bucket without falling below $-2$ in any other.
The cost and wall time served as secondary objectives.

% ------------------------------------------------------------------------------

\subsection{Phase~2: project-scale tasks}
\label{sec:evolution_phase2}

To prepare the second phase, we added file manipulation tools: \tool{open} (start an interactive session in a file), \tool{build} (compile a file), and a \tool{verify} tool that implements the anti-cheating protocol of Section~\ref{sec:cheating} for entire projects.
We also equip agents with file manipulation tools in a different interface: \tool{read}, \tool{write}, \tool{list} (list files).
Unlike other mutations, these tools are not evaluated or validated: we do not want the performance of \rme to be attributable to optimized file handling, or reward hacking.

The second evolution phase focuses on project-scale tasks.
Given the difficulty of the tasks, \testers are Sonnet instances with four runs per project, each limited to 900\,s of wall-clock time, with a safety cap of 200 server calls (never reached in practice).
\Testers receive no system prompt.
The server's own documentation is their only guidance.
We thus evaluate the tools and not the prompts.
We set a maximum budget of \$150 for Phase~2.

\textbf{Dataset}
The dataset used comprises five autoformalization projects described in natural language.
Each project requires the agents to generate definitions, functions, and lemmas.
The five projects span diverse topics in mathematics and computer science: (1) tropical algebra (min-plus), (2) closed real intervals, (3) an append-only ledger, (4) finite automata manipulation, and (5) a divisor-sum exercise (inspired by an IMO 2025 problem).
A description of each task is given in Table~\ref{tab:autoform_tasks} of the appendix.
We expect four to six files per project, each importing MathComp, the largest mathematical library in Rocq.
Every project ships with a test suite to check the correctness of the solution.

\textbf{Validation}
In Phase~2, evaluation comprises 20 runs in total (four runs for each of the five projects).
A mutation is validated if \testers gain more than two successes compared to the previous iteration.
A mutation is reverted if \testers lose more than two successes.
This threshold corresponds to the variation of $\pm 2$ observed with the \ctrlmcp on the five project-scale tasks.
When the difference is smaller than 3, a cost or wall time reduction of at least 20\,\% validates the mutation.

% ------------------------------------------------------------------------------

\subsection{Anti-cheating system}
\label{sec:cheating}

In both phases, we must ensure that the \testers' answers are correct.
Even with a proof assistant, agents sometimes try to cheat the experimental harness to report a success even if the task was not completed.
We thus designed an anti-cheating system.

For a proof, the system checks that the file prefix (including imports and statement) is left untouched, and that the code written by the agent contains no axioms, no partial proofs (e.g., using \texttt{Axiom}, \texttt{Parameter}, \texttt{admit}, \texttt{Admitted}, \texttt{Abort}, \ldots), and no additional imports (which can hide redefinitions).
The harness then recompiles the file alone in a clean directory and audits the theorem's assumptions with \texttt{Print Assumptions} against a list of standard-library axioms.
For a project, the system rebuilds the project in an isolated directory, compiles the test suite against it, and audits the assumptions on each test.

\section{\rme}
\label{sec:results}

\rme is entirely written in OCaml (the programming language used for the Rocq implementation) on top of the Rocq runtime API.
Implementation details can be found in Table~\ref{tab:footprint} of the appendix.
In line with existing IDEs for interactive theorem proving, the final server maintains a live interactive session with the agent.
The session captures the state of progress in the proof and does not need to be recomputed at each interaction, which is much more efficient in tokens and wall time than only exposing the compiler.
A call to Rocq drops from 266\,ms for a full compilation to about 1\,ms with the session mechanism.
\rme enables the agent to apply tactics to advance the proof state, to backtrack, and to access the current proof state.
The server also exposes simple automation tools (from the Rocq ecosystem) that can be used to close simple proof goals.
The server offers verification tools to check the validity of a proof or of an entire project based on the design described in Section~\ref{sec:cheating}.

\begin{figure}[t]
\centering
\input{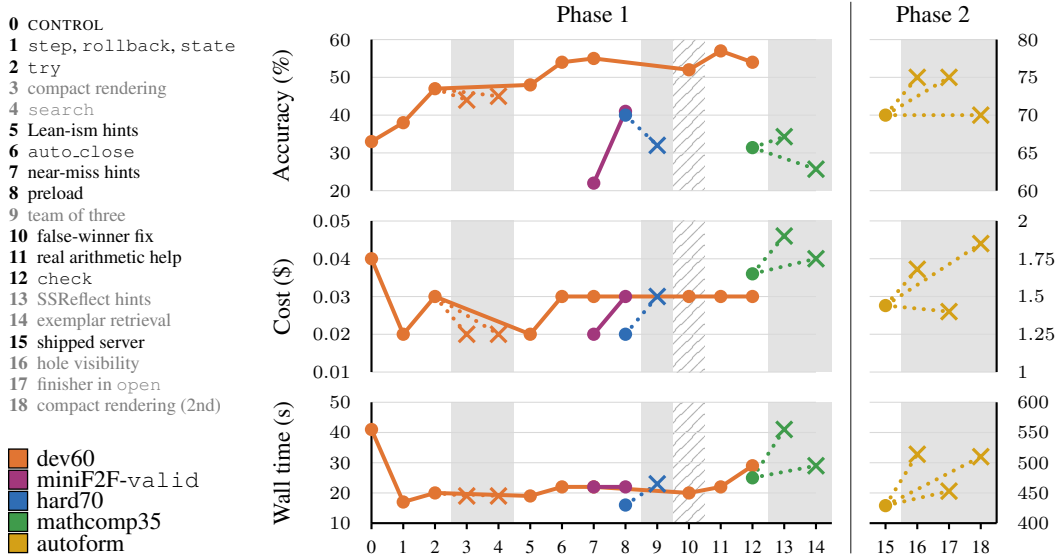}
\caption{Evolution of accuracy, cost, and wall time across both phases.
In the left column, the mutations are numbered chronologically and correspond to x-axis labels.
Numbers 0 and 15 are not mutations but the starting servers of the phases.
Colors indicate the dataset each point was measured on.
Each mutation is linked to the version it is compared to.
The link of a reverted mutation is dotted, its background is gray and its data point is a cross.
The background of a bug fix is hatched gray.
}
\label{fig:phase1_evolution}
\end{figure}

The evolutions of accuracy, cost, and wall time across mutations are presented in Figure~\ref{fig:phase1_evolution}.
During Phase~1, accuracy on the difficulty buckets of the dev60 dataset is the main driver and we observe that overall the accuracy increases with the accepted mutations, from $.33$ to $.54$.
To compare the final server to the naive \ctrlmcp we run a final test with four runs on dev60, i.e., 240 attempts.
Over the four runs, the final server solves more attempts than the naive \ctrlmcp on 22 problems, fewer on none.
During Phase~1, cost and wall time were not used to validate any mutation.
We observe that both metrics significantly improved in the final server compared to the naive \ctrlmcp.
The list of mutations along with their description is presented in Table~\ref{tab:evolution_chronology} of the appendix.
The detailed results of each mutation are given in Table~\ref{tab:evolution_deltas}.

\newcommand{\dd}[2]{${+}#1{-}#2$}

\begin{table}[t]
\caption{\textbf{Detailed results} of the evolution process. Each line corresponds to a mutation of Table~\ref{tab:evolution_chronology}. Each cell reports the change with respect to the previous version as a diff: the number of new successes ($+$) and failures ($-$). For each dataset we report the number of problems in each difficulty bucket, and the number of runs per problem. Phase~1 and Phase~2 are separated by bolder lines. Gray lines indicate reverted mutations. The hatched line is a bug fix (kept).}
\label{tab:evolution_deltas}
\begin{center}
\begin{small}
\renewcommand{\arraystretch}{1.2}
\setlength{\tabcolsep}{3pt}
\bugtabular{%
\begin{tabular}{>{\raggedright\arraybackslash}p{4.15cm}>{\raggedright\arraybackslash}p{4.45cm}cccc}
\bf Mutations & \bf Dataset & \bf Easy & \bf Medium & \bf Hard & \bf Total \\
\Xhline{1.2pt}

\keptrow \tool{step}, \tool{rollback}, \tool{state} & dev60, 20/20/20\,$\times$\,2 & \dd{4}{3} & \dd{4}{1} & \dd{4}{1} & \dd{12}{5} \\
\hline

\keptrow \tool{try} & dev60, 20/20/20\,$\times$\,2 & \dd{7}{0} & \dd{4}{2} & \dd{2}{0} & \dd{13}{2} \\
\hline

\revertedrow \gr{compact rendering} & \gr{dev60, 20/20/20\,$\times$\,2} & \gr{\dd{1}{2}} & \gr{\dd{1}{2}} & \gr{\dd{0}{1}} & \gr{\dd{2}{5}} \\
\hline

\revertedrow \gr{\tool{search}} & \gr{dev60, 20/20/20\,$\times$\,2} & \gr{\dd{0}{2}} & \gr{\dd{2}{3}} & \gr{\dd{2}{1}} & \gr{\dd{4}{6}} \\
\hline

\keptrow Lean-ism hints & dev60, 20/20/20\,$\times$\,2 & \dd{0}{2} & \dd{6}{2} & \dd{1}{1} & \dd{7}{5} \\
\hline

\keptrow \tool{auto\_close} & dev60, 20/20/20\,$\times$\,2 & \dd{2}{0} & \dd{5}{2} & \dd{3}{1} & \dd{10}{3} \\
\hline

\keptrow near-miss hints & dev60, 20/20/20\,$\times$\,2 & \dd{2}{0} & \dd{4}{2} & \dd{1}{1} & \dd{7}{3} \\
\hline

\keptrow preload & miniF2F-\texttt{valid}, 130/79/35\,$\times$\,2 & \dd{66}{2} & \dd{29}{3} & \dd{4}{0} & \dd{99}{5} \\
\hline

\revertedrow \gr{team of three} & \gr{hard70, 0/0/70\,$\times$\,2} & \gr{---} & \gr{---} & \gr{\dd{0}{11}} & \gr{\dd{0}{11}} \\
\hline

\bugrow false-winner fix & dev60, 20/20/20\,$\times$\,2 & \dd{0}{2} & \dd{1}{7} & \dd{2}{1} & \dd{3}{10} \\
\hline

\keptrow real arithmetic help & dev60, 20/20/20\,$\times$\,2 & \dd{1}{1} & \dd{7}{2} & \dd{2}{1} & \dd{10}{4} \\
\hline

\keptrow \tool{check}, Haiku & dev60, 20/20/20\,$\times$\,2 & \dd{1}{1} & \dd{2}{5} & \dd{1}{2} & \dd{4}{8} \\
\hline

\keptrow \tool{check}, Sonnet & dev60, 20/20/20\,$\times$\,2 & \dd{1}{0} & \dd{7}{0} & \dd{8}{2} & \dd{16}{2} \\
\hline

\revertedrow \gr{SSReflect hints} & \gr{mathcomp35, 35\,$\times$\,1} & \multicolumn{3}{c}{\gr{---}} & \gr{\dd{5}{4}} \\
\hline

\revertedrow \gr{exemplar retrieval} & \gr{mathcomp35, 35\,$\times$\,1} & \multicolumn{3}{c}{\gr{---}} & \gr{\dd{1}{3}} \\
\Xhline{1.2pt}

\keptrow \tool{open}, \tool{build}, \tool{verify} & --- & \multicolumn{4}{c}{\emph{not evaluated}} \\
\Xhline{1.2pt}

\revertedrow \gr{hole visibility} & \gr{autoform, 5 tasks\,$\times$\,4} & \multicolumn{3}{c}{\gr{---}} & \gr{\dd{2}{1}} \\
\hline

\revertedrow \gr{finisher in \tool{open}} & \gr{autoform, 5 tasks\,$\times$\,4} & \multicolumn{3}{c}{\gr{---}} & \gr{\dd{1}{0}} \\
\hline

\revertedrow \gr{compact rendering} & \gr{autoform, 5 tasks\,$\times$\,4} & \multicolumn{3}{c}{\gr{---}} & \gr{\dd{1}{1}} \\
\Xhline{1.2pt}

\end{tabular}}
\end{small}
\end{center}
\end{table}

Five mutations were not evaluated on dev60 alone.
The tool \tool{check} was introduced to mitigate a regression on Sonnet instances and was kept despite a modest loss on Haiku.
The preload mutation targets an issue that is not present in dev60.
Automation tactics (e.g., \texttt{lia} or \texttt{nra}) require an extra import that violates our anti-cheating protocol and the harness can reject valid proofs.
Preloading these tactics mitigates this issue, but such automation is not necessary on dev60.
This mutation was thus evaluated on the \texttt{valid} split of miniF2F-Rocq.
Agent collaboration (the ``team of three'' mutation) was evaluated on hard70 and rejected by the \orch.
Two MathComp-specific mutations (SSReflect hints and exemplar
retrieval) were evaluated once on mathcomp35 and rejected by the \orch.
The false-winner mutation is a bug fix: \tool{auto\_close} falsely reported a goal as closed when a tactic did not make progress.
The fix is listed in the mutations to separate its effect from the next iteration.

One of the most popular tools in existing MCP servers is \texttt{search} to search for applicable lemmas.
This tool was one of the first proposed mutations, but the evaluation revealed that while agents heavily used this tool, it yielded no gain in accuracy.

In Phase~2, three mutations were evaluated and rejected.
The \orch then concluded that the current server was at a local optimum for the project-scale tasks and aborted Phase~2 before exhausting its budget at ${\$116 / \$150}$.

\section{Evaluation}
\label{sec:evaluation}

% ------------------------------------------------------------------------------

To evaluate \rme, we compare it to two MCP servers across different model sizes and families: Claude Haiku~4.5, Claude Sonnet~5, Claude Opus~4.8, and GPT-5.6 Terra.
The baseline \ctrlmcp corresponds to the starting point of the evolution process and only exposes the Rocq compiler.
\rocqmcp is an established MCP server for Rocq~\citep{putnam2025,certigc-2026,agentic-tt-2026}.

We first evaluate the servers on the mathematical exercises of miniF2F-Rocq.
Then, we focus on three research questions:
Are tools evolved on exercises useful for building entire projects?
Do the benefits of the tools transfer to another proof assistant?
How much of the gains come from standard automation?

% ------------------------------------------------------------------------------

\subsection{miniF2F-Rocq}

The evaluation dataset is the \texttt{test} split of miniF2F-Rocq divided into the same difficulty buckets as the \texttt{valid} split in Section~\ref{sec:evolution} (130 in easy, 79 in medium, 35 in hard).
For each problem, agents have a total budget of 300\,s and 200 calls, and no system prompt (only the official documentation of the MCP servers).
We use 2 runs per problem.
Results for each of the two runs are reported in Table~\ref{tab:detailed_results_runs} of the appendix.
The three metrics measured are the same as in Section~\ref{sec:evolution}: accuracy, cost, and wall time.

Table~\ref{tab:detailed_results} presents the results for each model, server, and bucket.
\rme outperforms both the baseline and \rocqmcp in all difficulty buckets, and significantly improves overall performance compared to \ctrlmcp (at least $+30$ points on all models).
Over the two runs, \rme solves more attempts than \ctrlmcp on 91 to 104 problems, fewer on at most 4 (sign test over problems, $p < 10^{-21}$), and more attempts than \rocqmcp on 26 to 55 problems, fewer on at most 8 ($p < 3 \times 10^{-3}$).
We observe that the accuracy gap between \rme and \rocqmcp decreases with stronger models (from $+15$ points with Haiku, to $+8$ points with Sonnet, and $+4$ points with Opus).
However, when looking at the cost and wall time, we observe the opposite trend: better models benefit more from \rme than \ctrlmcp for both cost (from $-45\,\%$ with Haiku, to $-54\,\%$ with Sonnet, and $-59\,\%$ with Opus) and wall time (from $-63\,\%$ with Haiku, to $-64\,\%$ with Sonnet, and $-64\,\%$ with Opus).
\rme is also faster than \rocqmcp almost all the time (except the hard bucket with Sonnet), despite the fact that both servers expose interactive sessions.
Results on Terra are similar to those on Opus, which confirms these observations and demonstrates that the final MCP server is not overspecialized to a family of models.

\begin{table}[t]
\caption{Results on the \texttt{test} split of miniF2F: total (easy/medium/hard). For each model, the cost and wall time are only computed on attempts on the subset of problems solved by all three MCP servers in at least one run (results on the entire dataset are reported in Table~\ref{tab:detailed_results_full} of the appendix).
The size of this subset per difficulty bucket is given under the model name.}
\label{tab:detailed_results}
\begin{center}
\begin{small}
\renewcommand{\arraystretch}{1.2}
\small
\begin{tabular}{ll r@{\hspace{0.45em}}l r@{\hspace{0.45em}}l r@{\hspace{0.45em}}l}
\bf Models & \bf MCP servers & \multicolumn{2}{c}{\bf Accuracy} & \multicolumn{2}{c}{\bf Cost (\$)} & \multicolumn{2}{c}{\bf Wall time (s)} \\
\Xhline{1.2pt}

 & \ctrlmcp & .17 & \buck{.28}{.05}{.01} & .07 & \buck{.06}{.07}{.15} & 52 & \buck{50}{60}{86} \\
 & \rocqmcp & .33 & \buck{.54}{.11}{.07} & .05 & \buck{.05}{.05}{.06} & 32 & \buck{32}{30}{48} \\
\multirow{-3}{*}{\begin{tabular}[c]{@{}l@{}}\bf Haiku\\\buck{40}{4}{1}\end{tabular}} & \rme & \textbf{.48} & \buck{\textbf{.69}}{\textbf{.30}}{\textbf{.14}} & \textbf{.04} & \buck{\textbf{.04}}{\textbf{.03}}{\textbf{.05}} & \textbf{19} & \buck{\textbf{19}}{\textbf{16}}{\textbf{36}} \\
\hline

 & \ctrlmcp & .50 & \buck{.68}{.35}{.17} & .24 & \buck{.19}{.36}{.39} & 96 & \buck{76}{153}{153} \\
 & \rocqmcp & .72 & \buck{.88}{.53}{.56} & .16 & \buck{.12}{.28}{.23} & 44 & \buck{30}{86}{\textbf{64}} \\
\multirow{-3}{*}{\begin{tabular}[c]{@{}l@{}}\bf Sonnet\\\buck{90}{28}{7}\end{tabular}} & \rme & \textbf{.80} & \buck{\textbf{.91}}{\textbf{.68}}{\textbf{.67}} & \textbf{.11} & \buck{\textbf{.08}}{\textbf{.18}}{\textbf{.22}} & \textbf{35} & \buck{\textbf{23}}{\textbf{62}}{77} \\
\hline

 & \ctrlmcp & .43 & \buck{.60}{.28}{.13} & .28 & \buck{.21}{.47}{.68} & 93 & \buck{67}{163}{206} \\
 & \rocqmcp & .72 & \buck{.88}{.56}{.46} & .19 & \buck{.14}{.34}{.35} & 42 & \buck{28}{83}{81} \\
\multirow{-3}{*}{\begin{tabular}[c]{@{}l@{}}\bf Opus\\\buck{83}{24}{5}\end{tabular}} & \rme & \textbf{.76} & \buck{\textbf{.90}}{\textbf{.61}}{\textbf{.54}} & \textbf{.12} & \buck{\textbf{.08}}{\textbf{.22}}{\textbf{.24}} & \textbf{33} & \buck{\textbf{20}}{\textbf{73}}{\textbf{67}} \\
\hline

 & \ctrlmcp & .54 & \buck{.70}{.40}{.27} & .06 & \buck{.04}{.08}{.10} & 84 & \buck{70}{107}{135} \\
 & \rocqmcp & .80 & \buck{.92}{.66}{.63} & .03 & \buck{.02}{.05}{.05} & 42 & \buck{29}{74}{63} \\
\multirow{-3}{*}{\begin{tabular}[c]{@{}l@{}}\bf Terra\\\buck{99}{33}{12}\end{tabular}} & \rme & \textbf{.88} & \buck{\textbf{.97}}{\textbf{.80}}{\textbf{.71}} & \textbf{.02} & \buck{\textbf{.01}}{\textbf{.03}}{\textbf{.04}} & \textbf{34} & \buck{\textbf{23}}{\textbf{58}}{\textbf{60}} \\
\Xhline{1.2pt}

\end{tabular}
\end{small}
\end{center}
\end{table}

On the subset of problems solved by all three MCP servers in at least one run, \rme improves cost and wall time per solve in almost all difficulty buckets.
This result may be surprising considering that the evolution process was only driven by the per-bucket accuracy.
This improvement is a consequence of the experimental harness used in Phase~1.
The 300\,s and 30-call budget forced the \orch to optimize the compilation time and the number of generated tokens by introducing the interactive session, then by maximizing the density of information returned by the tools (hints in error messages from \tool{try} and \tool{auto\_close}).
The numbers of calls and tokens generated across models are presented in Table~\ref{tab:efficiency_results}.
These results are computed on problems solved with all three servers per model and averaged over the four models.
We observe that the number of calls of \rme is the lowest among the three MCP servers.
\rme lowers the number of input tokens by 24\,\% and divides by four the number of output tokens compared to the \ctrlmcp server.
Fewer tokens and tool calls then directly translate to lower cost and wall time.

% ------------------------------------------------------------------------------

\subsection{RQ1: Are tools evolved on exercises useful for building entire projects?}
\label{sec:RQ1}

In Phase~2, the \orch proposed three mutations that were all rejected.
Since no mutation was accepted on the project-scale tasks, we reuse these tasks here to evaluate the final server of Phase~1.
This dataset was used during the evolution, but the server did not adapt to these tasks, so we can evaluate whether the exposed tools are useful to build entire projects.

We evaluate three models (Sonnet, Opus, and Terra) on the five project-scale tasks of Phase~2 (Section~\ref{sec:evolution_phase2}, Table~\ref{tab:autoform_tasks}).
Given the difficulty of the tasks, we did not try with the smallest model, Haiku.
For each project, agents have a total budget of 900\,s and 200 calls.
In addition to the three MCP servers, we expose the following common tools: \tool{write}, \tool{read}, \tool{list}, \tool{build}, and \tool{verify}.
We use four runs per task for Sonnet and Terra but only two for Opus (to limit the costs).
A run is a success if the project can be built from scratch, passes the test suite, and does not violate the anti-cheating protocol of Section~\ref{sec:cheating}.
Cost and wall time are averaged over solved runs.
Standard deviations across runs are reported in Table~\ref{tab:autoform_results_sd} of the appendix.

\begin{table}[t]
\caption{Efficiency results for the different MCP servers, averaged over the four models.}
\label{tab:efficiency_results}
\begin{center}
\begin{small}
\renewcommand{\arraystretch}{1.2}
\small
\begin{tabular}{lrrrr}
\bf MCP servers & \bf Calls & \bf Input tokens & \bf Output tokens & \bf Output tokens per call \\
\Xhline{1.2pt}

\ctrlmcp & 5.9 & 79.0k & 6.6k & 1.12k \\
\rocqmcp & 7.2 & 148.4k & 2.7k & 0.38k \\
\rme & \bf 5.6 & \bf 59.9k & \bf 1.6k & \bf 0.29k \\
\Xhline{1.2pt}

\end{tabular}
\end{small}
\end{center}
\end{table}

\begin{table}[t]
\caption{Results on project-scale tasks for the models Sonnet\,$\mid$\,Opus\,$\mid$\,Terra averaged on all tasks (detailed results are reported in Table~\ref{tab:autoform_results_full} of the appendix).}
\label{tab:autoform_results}
\begin{center}
\begin{small}
\renewcommand{\arraystretch}{1.2}
\small
\begin{tabular}{l@{\grpsep}r@{\midsep}r@{\midsep}r@{\grpsep}r@{\midsep}r@{\midsep}r@{\grpsep}r@{\midsep}r@{\midsep}r}
% A \multicolumn span swallows the group separator that follows it, so each
% label is padded by that same width to stay centred over its own numbers.
% \kern, not \hspace: array's cell template ends with \unskip, which drops
% trailing glue but leaves a kern alone.
\bf MCP servers & \multicolumn{3}{@{}c@{}}{\bf Accuracy\kern\grpsepwidth} & \multicolumn{3}{@{}c@{}}{\bf Cost (\$)\kern\grpsepwidth} & \multicolumn{3}{@{}c@{}}{\bf Wall time (s)\kern\tabcolsep} \\
\Xhline{1.2pt}

\ctrlmcp & .50 & .30 & .40 & 1.56 & 2.19 & 0.22 & 608 & 714 & 332 \\
\rocqmcp & .60 & \bf .60 & .35 & 2.63 & 2.48 & 0.34 & 595 & 661 & 502 \\
\rme & \bf .70 & \bf .60 & \bf .50 & \bf 1.44 & \bf 1.89 & \bf 0.16 & \bf 429 & \bf 574 & \bf 269 \\
\Xhline{1.2pt}

\end{tabular}
\end{small}
\end{center}
\end{table}

Results are presented in Table~\ref{tab:autoform_results} averaged on all tasks.
Again, we observe that for all three metrics, accuracy, cost, and wall time, models equipped with \rme yield the best performance.
Compared to \ctrlmcp, the accuracy gain is $+20$ points with Sonnet ($14/20$ against $10/20$ successes), $+30$ points with Opus ($6/10$ against $3/10$), and $+10$ points with Terra ($10/20$ against $8/20$).
\rocqmcp also improves the performance of the Claude family models, and is on par with \rme with Opus, but is less performant than \ctrlmcp with Terra ($7/20$ against $8/20$).
The cost and wall time trends are similar to the miniF2F-Rocq evaluation with better gain for stronger models both on cost ($-8\,\%$ with Sonnet, $-14\,\%$ with Opus, and $-27\,\%$ with Terra) and wall time ($-29\,\%$ with Sonnet, $-20\,\%$ with Opus, and $-19\,\%$ with Terra).
\rocqmcp is the most expensive server, and it is slower than \rme with all models.

% ------------------------------------------------------------------------------

\subsection{RQ2: Do results transfer to another proof assistant?}

To test whether the results of \rme transfer to Lean, another proof assistant, we build \lme.
\lme is a port to Lean of \rme, with the exact same set of tools.
We compare this server to \leanmcp~\citep{lean-lsp-mcp}, a popular MCP server for Lean, and \ctrlmcp, a simple server that only exposes the compiler (via \texttt{lake build}).

The evaluation dataset putnam60 comprises 60 problems from PutnamBench~\citep{putnambench}.
We selected 20 easy, 20 medium, and 20 hard problems.
To assess the difficulty of a problem, we used the number of published LLM-based methods that report a solution (see Table~\ref{tab:putnam_selected_models} in the appendix).
In this experiment we only use Sonnet with two runs per problem.
Results for each of the two runs are reported in Table~\ref{tab:lean_results_runs} of the appendix.
For each problem, agents have a total budget of 300\,s and 200 calls.

Table~\ref{tab:lean_results} presents the results with Sonnet.
\leanmcp outperforms both \ctrlmcp and \lme on accuracy, but \lme yields the best cost and wall time.
In contrast with Rocq, \lme and \ctrlmcp give the same accuracy.
As explained in Section~\ref{sec:intro}, we cannot rule out contamination of Sonnet on PutnamBench (many Lean solutions are now available on GitHub), and we cannot rule out that Sonnet was trained on \leanmcp (like Leanstral~\citep{leanstral}).

\begin{table}[t]
\caption{Results on putnam60 with Sonnet in Lean for the three MCP servers.
Cost and wall time are averaged over problems solved by all three MCP servers in at least one run (16 easy problems, 2 medium problems, and no hard problems).}

\label{tab:lean_results}
\begin{center}
\begin{small}
\renewcommand{\arraystretch}{1.2}
\small
\begin{tabular}{l r@{\hspace{0.45em}}l r@{\hspace{0.45em}}l r@{\hspace{0.45em}}l}
\bf MCP servers & \multicolumn{2}{c}{\bf Accuracy} & \multicolumn{2}{c}{\bf Cost (\$)} & \multicolumn{2}{c}{\bf Wall time (s)} \\
\Xhline{1.2pt}

\ctrlmcp & .33 & \buck{.93}{.05}{.00} & .12 & \buck{.11}{\textbf{.28}}{--} & 82 & \buck{76}{183}{--} \\
\leanmcp & \textbf{.42} & \buck{\textbf{.95}}{\textbf{.30}}{.00} & .14 & \buck{.11}{.40}{--} & 86 & \buck{72}{199}{--} \\
\lme & .33 & \buck{.83}{.18}{.00} & \textbf{.11} & \buck{\textbf{.09}}{.35}{--} & \textbf{64} & \buck{\textbf{55}}{\textbf{159}}{--} \\
\Xhline{1.2pt}

\end{tabular}
\end{small}
\end{center}
\end{table}

% ------------------------------------------------------------------------------

\subsection{RQ3: How much of the gains come from standard automation?}

Automation is a powerful tool in proof assistants and a carefully crafted tactic can solve a lot of problems without any interaction with the user (or agent).
The addition of \tool{auto\_close}, which tries a portfolio of automatically closing tactics, yielded a noticeable accuracy gain during the evolution (mutation 6 in Figure~\ref{fig:phase1_evolution}).
To quantify how many successes are due to this tool compared to agent interactions, we run \tool{auto\_close} on all the problems of the \texttt{test} split of miniF2F-Rocq.
The tool solved $52/244$ problems, i.e., $21\,\%$ ($.35$\,/\,$.06$\,/\,$.03$ per bucket).
On this subset of problems, the cost and wall time with \rme averaged over all models are \$0.03 and 11\,s respectively, against \$0.06 and 19\,s with \rocqmcp.
But the gains are mostly in the easy bucket, and automation alone does not explain the performance of \rme compared to \ctrlmcp and \rocqmcp.

\section{Related work}
\label{sec:related}

\textbf{Agent/Prover interfaces}
There exist multiple MCP servers and REPL interfaces to connect agents with the Lean theorem prover~\citep{leandojo,pantograph,seedprover,axle,numina-lean-agent}.
These interfaces differ in the set of tools exposed to the agent, and most of the work focuses on speed, robustness, and scalability.
To the best of our knowledge, none of these interfaces followed an evolutionary design process to optimize the set of tools.
\rocqmcp was designed by a frontier model (Claude Opus 4.6) by analyzing logs from a prior experiment~\citep{putnam2025} but falls short of measuring the added value of each tool, is limited to one iteration, and was never rigorously evaluated.

\textbf{Evolving interfaces}
In SWE-agent~\citep{yang2024sweagent} the agent interface is already studied experimentally, and multiple works evaluate the impact of different tools on agent capabilities~\citep{xu2026devil,mak2026bash}.
Several lines of work then follow an iterative process to develop agentic frameworks starting with agents tasked to build reusable tools~\citep{cai2024latm,wang2024voyager,wang2024trove}, or MCP tools implemented at runtime~\citep{qiu2025alita}.
Instead of building a library of task-specific tools, we try to optimize a small set of specialized tools for the prover.
Another line of work focuses on evolving the entire agentic harness with self-modifying coding agents~\citep{zhang2026dgm,robeyns2025sica}, tool evolution~\citep{lin2026ahe}, or harness search~\citep{lee2026metaharness}.
Our process evolves one feature at each step, and we show that the gains on three metrics hold across different sizes and families of models.
The closest work focuses on self-modifying Lean proof agents~\citep{li2026selfmodifying} where an evolutionary process optimizes a proof-repair pipeline.
But evolved tools mostly check lemma names to avoid hallucinations, and the evaluation focuses on standalone theorems from miniF2F.
We focus on tools that can handle project-scale tasks.
Specialized for the Rocq prover, RocqSmith focuses on optimizing prompts and control flow~\citep{kozyrev2026rocqsmith}.

\section{Conclusion and future work}

We presented an evolutionary method to grow an agent/prover interface: a frontier model proposes one feature at a time, and a feature is kept only if it measurably improves the accuracy of smaller models on a fixed set of problems, with cost and wall time as secondary objectives.
This process grew \rme from a server that only exposes the Rocq compiler.
On the held-out \texttt{test} split of miniF2F-Rocq, \rme outperforms both the \ctrlmcp baseline and the established \rocqmcp server in success rate, cost per solve, and time per solve, across four models from two families, and the same ordering holds on project-scale autoformalization tasks.
Although only accuracy drove the validation of mutations, the tight time and call budgets of the \testers led the \orch to reduce the number of calls and tokens, which translated into lower cost and wall time on every model we tested.

The natural next step is to run the same procedure for Lean.
Our port of the evolved tools to Lean, \lme, already improves cost and time per solve on a subset of PutnamBench~\citep{putnambench}, but its solve rate remains below \leanmcp~\citep{lean-lsp-mcp}.
Growing a Lean server from a compiler-only \ctrlmcp, with the same objectives and validation rules, would let the \orch discover Lean-specific features.
It would also tell which features of \rme are intrinsic to agent/prover interaction and which are specific to Rocq.
%Beyond Lean, the process is agnostic to the prover and to the objectives: the same loop could target other metrics, such as robustness on long projects, or use the evolved interface as a training environment for smaller models.

\clearpage

\subsection*{AI use statement}

Generative AI is both the object of study and a tool in this work; we distinguish the two roles below.

\textbf{Required disclosures}
We used generative AI tools to implement methods, to design and evaluate experiments, to propose and refine hypotheses, and to interpret results.
As described in Section~\ref{sec:evolution}, the \orch (Claude Fable~5) chose the time, call, and cost budgets of the experiments, proposed every mutation of the server, implemented it, chose the dataset on which to evaluate it, interpreted the results of the \testers against the validation rules we fixed in advance, and decided to keep or revert it, including the decision to end Phase~2 before exhausting its budget.
The \testers (Claude Haiku~4.5 and Claude Sonnet~5) and the models of the evaluation (adding Claude Opus~4.8 and GPT-5.6 Terra) are the experimental subjects; the Rocq and Lean proofs they produce are measured outputs.
We also used generative AI tools to generate a synthetic dataset: the five autoformalization projects of Phase~2, with their natural-language specifications, reference solutions, and test suites, were produced by the \orch and reviewed by the authors.
We also used generative AI tools to clean and reformat datasets: the selection of dev60, hard70, and mathcomp35 and the difficulty-bucketing heuristic were implemented by the \orch.
We have not used generative AI tools to develop the conceptual framework of this work: the evolutionary method, the three objectives, the validation rules, and the anti-cheating protocol were designed by the authors; the \orch then implemented the validation rules and the anti-cheating protocol.
Formulating mathematical claims, providing ingredients for proofs, assisting in the writing of proofs, translation, and qualitative or thematic data analysis are not applicable: this paper contains no mathematical claims of its own.

\textbf{Recommended disclosures}
Additionally, we used generative AI tools to create and edit software code (the \rme server, its port to Lean, the evaluation harness, and the scripts that produce the tables and figures from the run records), to draft parts of this paper (the \orch wrote a technical report of the experiment from its own logs, which we used as a source of facts alongside the raw logs; the present manuscript was written by the authors, with a coding agent used for the first draft of this statement), to edit the paper for readability, and to search for and summarize related literature.

\textbf{Review}
We have reviewed all AI-assisted work.
The evolution ran under objectives and validation rules fixed by the authors before it started, and the authors reviewed the decision trail of the \orch daily.
No output reported by an agent or by its tools is trusted: every proof and project counted as a success was recompiled from scratch in a clean directory and audited with \texttt{Print Assumptions} (Section~\ref{sec:cheating}), and the final server was frozen before the held-out \texttt{test} split of miniF2F-Rocq was touched.
The numbers in the tables were recomputed by the authors from the raw run records.
AI-generated code is released with the paper.
Every claim in the text was checked by the authors against the logs, and every reference was checked against the original publication.
We take responsibility for the final content of this work, including text, claims, or artifacts produced with the aid of generative AI.

% \subsection*{Ethics statement}

% (This section is \textbf{recommended} and does not count toward the page limit.)

% If authors feel that their paper submission raises questions regarding the Code
% of Ethics, they are encouraged to include a paragraph of Ethics Statement (at
% the end of the main text before references) to address potential concerns where
% appropriate. Topics include, but are not limited to, studies that involve human
% subjects, practices to data set releases, potentially harmful insights,
% methodologies and applications, potential conflicts of interest and sponsorship,
% discrimination/bias/fairness concerns, privacy and security issues, legal
% compliance, and research integrity issues (e.g., IRB, documentation, research
% ethics). This statement should not be more than 1 page.

\subsection*{Reproducibility statement}

% CHECK (authors): add the anonymized repository link for submission, and confirm whether the raw run records and the decision trail of the orchestrator are released with the code.
We release the code of the Rocq experiments: the \rme server, the \ctrlmcp server, the evaluation harness with the anti-cheating verifier of Section~\ref{sec:cheating}, the problem selections (dev60, hard70, mathcomp35, and the difficulty buckets of miniF2F-Rocq), the five autoformalization projects of Table~\ref{tab:autoform_tasks} with their test suites, and the scripts that produce every table and figure from the run records.
All of the above is available at an anonymized repository: \url{https://osf.io/648ex/overview?view_only=b6294376221b4dd1935b604f571abbce}.
We will also release the \lme code and experiments upon acceptance.
The evolutionary process is described in Section~\ref{sec:evolution}: the two phases, their datasets, the budgets of the \testers, and the rules that accept or revert a mutation.
Since the process is driven by a frontier model, its outcome cannot be replayed deterministically, but it can be audited: Tables~\ref{tab:evolution_chronology} and~\ref{tab:evolution_deltas} list every proposed mutation, the dataset on which it was evaluated, its per-bucket results, and its verdict, and the released server is the frozen result of this chronology.
The evaluation protocol (models, budgets, absence of system prompt, difficulty buckets, and success criterion) is given in Section~\ref{sec:evaluation}, with per-model and per-bucket results in Table~\ref{tab:detailed_results} and complete results in Tables~\ref{tab:detailed_results_full} and~\ref{tab:autoform_results_full} of the appendix.
All agents are proprietary models accessed through their public APIs; we name their exact versions, and we report cost in US dollars at the prices in force at the time of the experiments and wall time as measured on our infrastructure, both of which may drift with future model or pricing updates.
% CHECK (authors): the miniF2F test setup in the evaluation section does not state its number of runs per problem, nor whether accuracy is averaged over runs; add it there.
To limit the effect of sampling noise, most results aggregate two to four runs per problem, the number of runs is given with each experiment, and cost and wall time are compared on the attempts solved by all servers.

\subsection*{Acknowledgements}

We thank Laetitia Teorodescu for the discussion, during a taxi ride back from Oléron, that led to this experiment.
This work was partially supported by the Défi Inria LLM4Code, the Groupe Casino/ENS Chair on Algorithmics and Machine Learning, and the Renaissance Philanthropy project “Leaning and Rocq’ing”.
This work was granted access to the HPC resources of IDRIS under the allocation A0191016874 made by GENCI.

% \subsubsection*{Author Contributions}
% If you'd like to, you may include  a section for author contributions as is done
% in many journals. This is optional and at the discretion of the authors.

% \subsubsection*{Acknowledgments}
% Use unnumbered third level headings for the acknowledgments. All
% acknowledgments, including those to funding agencies, go at the end of the paper.

\clearpage

\bibliography{iclr2027_conference}
\bibliographystyle{iclr2027_conference}

\clearpage
\appendix
% The reduced float spacing (\floatgap) only applies to the main text: back to
% the LaTeX defaults in the appendix.
\setlength{\textfloatsep}{20pt plus 2pt minus 4pt}
\setlength{\floatsep}{12pt plus 2pt minus 2pt}
\setlength{\intextsep}{12pt plus 2pt minus 2pt}
\section{Appendix}

\begin{table}[h]
\caption{The five project-scale tasks. Each task requires the creation of multiple files on top of the MathComp library.}
\label{tab:autoform_tasks}
\begin{center}
\begin{small}
\renewcommand{\arraystretch}{1.2}
\begin{tabular}{lcp{9.4cm}}
\bf Task & \bf Files & \bf Description \\
\Xhline{1.2pt}
\texttt{frugal} & 4 & bounded min-plus cost algebra on \texttt{option nat} with a $2\times2$ matrix product; laws of the operations and associativity of the product \\
\hline
\texttt{gauges} & 4 & closed real intervals over an abstract \texttt{realType} with a sound interval arithmetic; three soundness lemmas and three width lemmas \\
\hline
\texttt{ledger} & 4 & append-only ledger of signed deltas with replay and checkpoints; replay/checkpoint equivalence and a bound on balances \\
\hline
\texttt{prodauto} & 6 & deterministic finite automata over a finite type; product and complement constructions, language intersection, emptiness by bounded reachability, a pigeonhole pumping lemma \\
\hline
\texttt{triadic} & 4 & a divisor-combinatorics dynamical system on natural numbers (from an IMO 2025 problem); a fixed-point theorem and a shrinking theorem \\
\Xhline{1.2pt}
\end{tabular}
\end{small}
\end{center}
\end{table}

\begin{table}[h]
\caption{\textbf{Implementation}. \rocqmcp is implemented on top of a mature software stack developed for IDE support and exposes capabilities this benchmark does not exercise (file outlines, notation resolution, operational diagnostics). \rme is implemented on top of the Rocq API.}
\label{tab:footprint}
\begin{center}
\begin{small}
\renewcommand{\arraystretch}{1.2}
\setlength{\tabcolsep}{4pt}
\begin{tabular}{lll}
& \bf \rme & \bf \rocqmcp \\
\Xhline{1.2pt}

implementation & OCaml & Python \\
first-party source & 3\,527 lines (7 files) & 7\,351 lines (8 files) \\
installed components & server + multi-agent daemon/shim & server \\
direct dependencies & 4 opam packages & 3 PyPI packages + coq-lsp toolchain \\
prover attachment & rocq-runtime linked in-process & petanque (coq-lsp) protocol \\
runtime topology & one OS process & server + petanque subprocess; coqc per compile \\
shipped user docs & 2.4k words & 4.7k words \\
\Xhline{1.2pt}

\end{tabular}
\end{small}
\end{center}
\end{table}

\newcommand{\kept}{$\checkmark$}
\newcommand{\neither}{$-$}
\newcommand{\reverted}{$\times$}

\begin{table}[h]
\caption{\textbf{Mutations} proposed during the evolution process. Phase~1, the inter-phase, and Phase~2 are separated by bolder lines. Gray lines indicate reverted mutations. The hatched line is a bug fix (kept). \kept{} accepted, \reverted{} reverted, \neither{} added without evaluation. The exposed tools are written in this \tool{font}.}
\label{tab:evolution_chronology}
\begin{center}
\begin{small}
\renewcommand{\arraystretch}{1.2}
\setlength{\tabcolsep}{4pt}
\bugtabular{%
\begin{tabular}{llc}
\bf Mutations & \bf Descriptions & \bf Verdicts \\
\Xhline{1.2pt}

\keptrow \tool{step}, & execute sentences, commit good prefixes and report failures & \\
\keptrow \tool{rollback}, & swap to a previous snapshot in constant time & \\
\keptrow \tool{state} & render the open goals and the committed script & \multirow{-3}{*}{\kept} \\
\hline

\keptrow \tool{try} & test up to eight candidates and commit the first success & \kept \\
\hline

\revertedrow \gr{compact rendering} & \gr{goals printed with only hypotheses deltas to save tokens} & \gr{\reverted} \\
\hline

\revertedrow \gr{\tool{search}} & \gr{execute Rocq's \texttt{Search}/\texttt{About} commands over a loaded library} & \gr{\reverted} \\
\hline

\keptrow & rewrite Lean-like syntax and common syntax mistakes into & \\
\keptrow \multirow{-2}{*}{Lean-ism hints} & Rocq form in error messages & \multirow{-2}{*}{\kept} \\
\hline

\keptrow \tool{auto\_close} & run a portfolio of Rocq automatic tactics & \kept \\
\hline

\keptrow near-miss hints & append near actual lemma names to unknown-reference errors & \kept \\
\hline

\keptrow preload & preload Lia/Lra/Psatz, refuse mid-proof imports & \kept \\
\hline

\revertedrow \gr{team of three} & \gr{a coordinator, a worker, and a finisher sharing a proof session} & \gr{\reverted} \\
\hline

\bugrow false-winner fix & bug fix in the \tool{auto\_close} tool & \kept \\
\hline

\keptrow real arithmetic help & introduce \texttt{0 <= t²} hypotheses in the context for \tool{auto\_close} & \kept \\
\hline

\keptrow & whole-proof check, & \\
\keptrow & additional \texttt{Qed}-gate for proof completion in the server, & \\
\keptrow \multirow{-3}{*}{\tool{check}} & penalizing tactics in the \tool{auto\_close} portfolio removed & \multirow{-3}{*}{\kept} \\
\hline

\revertedrow \gr{SSReflect hints} & \gr{MathComp-specific hints} & \gr{\reverted} \\
\hline

\revertedrow \gr{exemplar retrieval} & \gr{provide similar proved lemmas from the project at start} & \gr{\reverted} \\
\Xhline{1.2pt}

\keptrow \tool{open} & open a proof session for the given file and optional theorem & \neither \\
\hline

\keptrow \tool{build} & compile a Rocq file and return detailed results & \neither \\
\hline

\keptrow \tool{verify} & verify the correctness of a whole project directory & \neither \\
\Xhline{1.2pt}

\revertedrow \gr{hole visibility} & \gr{all tools warn about remaining `admit.` in the proof} & \gr{\reverted} \\
\hline

\revertedrow \gr{finisher in \tool{open}} & \gr{\tool{open} runs the \tool{auto\_close} portfolio automatically} & \gr{\reverted} \\
\hline

\revertedrow \gr{compact rendering} & \gr{try again the compact rendering of Phase~1} & \gr{\reverted} \\
\Xhline{1.2pt}
\end{tabular}}
\end{small}
\end{center}
\end{table}

\begin{table}[h]
\caption{Results on the \texttt{test} split of miniF2F-Rocq with cost and wall time computed on the whole dataset instead of the problems solved with all three MCP servers.}
\label{tab:detailed_results_full}
\begin{center}
\begin{small}
\renewcommand{\arraystretch}{1.2}
\begin{tabular}{ll r@{\hspace{0.45em}}l r@{\hspace{0.45em}}l r@{\hspace{0.45em}}l}
\bf Models & \bf MCP servers & \multicolumn{2}{c}{\bf Accuracy} & \multicolumn{2}{c}{\bf Cost (\$)} & \multicolumn{2}{c}{\bf Wall time (s)} \\
\Xhline{1.2pt}

 & \ctrlmcp & .17 & \buck{.28}{.05}{.01} & \textbf{.07} & \buck{\textbf{.07}}{\textbf{.07}}{\textbf{.15}} & 54 & \buck{53}{\textbf{60}}{\textbf{86}} \\
\bf Haiku & \rocqmcp & .33 & \buck{.54}{.11}{.07} & .10 & \buck{.09}{.12}{.17} & 54 & \buck{51}{66}{96} \\
 & \rme & \textbf{.48} & \buck{\textbf{.69}}{\textbf{.30}}{\textbf{.14}} & .13 & \buck{.10}{.24}{.23} & 67 & \buck{51}{116}{136} \\
\hline

 & \ctrlmcp & .50 & \buck{.68}{.35}{.17} & .25 & \buck{.20}{.38}{.39} & 100 & \buck{77}{161}{153} \\
\bf Sonnet & \rocqmcp & .72 & \buck{.88}{.53}{.56} & .25 & \buck{.17}{.36}{.46} & 73 & \buck{47}{\textbf{114}}{142} \\
 & \rme & \textbf{.80} & \buck{\textbf{.91}}{\textbf{.68}}{\textbf{.67}} & \textbf{.21} & \buck{\textbf{.13}}{\textbf{.33}}{\textbf{.37}} & \textbf{72} & \buck{\textbf{41}}{115}{\textbf{127}} \\
\hline

 & \ctrlmcp & .43 & \buck{.60}{.28}{.13} & .29 & \buck{.21}{.49}{.68} & 95 & \buck{67}{169}{206} \\
\bf Opus & \rocqmcp & .72 & \buck{.88}{.56}{.46} & .34 & \buck{.24}{.48}{.61} & 83 & \buck{57}{121}{\textbf{163}} \\
 & \rme & \textbf{.76} & \buck{\textbf{.90}}{\textbf{.61}}{\textbf{.54}} & \textbf{.25} & \buck{\textbf{.16}}{\textbf{.36}}{\textbf{.55}} & \textbf{79} & \buck{\textbf{49}}{\textbf{116}}{165} \\
\hline

 & \ctrlmcp & .54 & \buck{.70}{.40}{.27} & .06 & \buck{.04}{.08}{.10} & 85 & \buck{71}{111}{135} \\
\bf Terra & \rocqmcp & .80 & \buck{.92}{.66}{.63} & .05 & \buck{.03}{.08}{.08} & 66 & \buck{44}{102}{101} \\
 & \rme & \textbf{.88} & \buck{\textbf{.97}}{\textbf{.80}}{\textbf{.71}} & \textbf{.04} & \buck{\textbf{.02}}{\textbf{.06}}{\textbf{.06}} & \textbf{62} & \buck{\textbf{41}}{\textbf{93}}{\textbf{93}} \\
\Xhline{1.2pt}

\end{tabular}
\end{small}
\end{center}
\end{table}

\begin{table}[h]
\caption{Results on the \texttt{test} split of miniF2F-Rocq for each of the two runs. Same conventions as Table~\ref{tab:detailed_results}: cost and wall time are computed on the problems solved by all three MCP servers.}
\label{tab:detailed_results_runs}
\begin{center}
\begin{small}
\renewcommand{\arraystretch}{1.2}
\begin{tabular}{llc r@{\hspace{0.45em}}l r@{\hspace{0.45em}}l r@{\hspace{0.45em}}l}
\bf Models & \bf MCP servers & \bf Run & \multicolumn{2}{c}{\bf Accuracy} & \multicolumn{2}{c}{\bf Cost (\$)} & \multicolumn{2}{c}{\bf Wall time (s)} \\
\Xhline{1.2pt}

 & \ctrlmcp & 1 & .17 & \buck{.28}{.05}{.03} & .06 & \buck{.06}{.07}{.15} & 51 & \buck{49}{64}{86} \\
 &  & 2 & .17 & \buck{.29}{.05}{.00} & .07 & \buck{.07}{.08}{--} & 52 & \buck{52}{56}{--} \\
 & \rocqmcp & 1 & .34 & \buck{.55}{.10}{.09} & .05 & \buck{.05}{.04}{.05} & 32 & \buck{32}{33}{40} \\
 &  & 2 & .33 & \buck{.53}{.11}{.06} & .06 & \buck{.06}{.05}{.07} & 33 & \buck{32}{28}{56} \\
 & \rme & 1 & .48 & \buck{.69}{.28}{.14} & .03 & \buck{.03}{.03}{.04} & 19 & \buck{19}{17}{28} \\
\multirow{-6}{*}{\bf Haiku} &  & 2 & .49 & \buck{.68}{.32}{.14} & .04 & \buck{.04}{.03}{.06} & 20 & \buck{19}{15}{43} \\
\hline

 & \ctrlmcp & 1 & .50 & \buck{.67}{.34}{.20} & .24 & \buck{.18}{.36}{.44} & 94 & \buck{71}{150}{169} \\
 &  & 2 & .50 & \buck{.69}{.35}{.14} & .24 & \buck{.20}{.37}{.33} & 99 & \buck{80}{156}{131} \\
 & \rocqmcp & 1 & .73 & \buck{.88}{.54}{.63} & .16 & \buck{.12}{.28}{.27} & 44 & \buck{30}{83}{78} \\
 &  & 2 & .70 & \buck{.88}{.51}{.49} & .16 & \buck{.12}{.28}{.19} & 45 & \buck{31}{89}{50} \\
 & \rme & 1 & .82 & \buck{.92}{.71}{.69} & .11 & \buck{.08}{.20}{.23} & 36 & \buck{22}{70}{81} \\
\multirow{-6}{*}{\bf Sonnet} &  & 2 & .79 & \buck{.90}{.66}{.66} & .11 & \buck{.08}{.17}{.20} & 33 & \buck{24}{53}{73} \\
\hline

 & \ctrlmcp & 1 & .44 & \buck{.62}{.29}{.11} & .30 & \buck{.24}{.49}{.62} & 98 & \buck{74}{168}{186} \\
 &  & 2 & .43 & \buck{.59}{.28}{.14} & .26 & \buck{.18}{.46}{.73} & 88 & \buck{60}{159}{222} \\
 & \rocqmcp & 1 & .73 & \buck{.89}{.54}{.51} & .19 & \buck{.14}{.35}{.36} & 44 & \buck{29}{85}{89} \\
 &  & 2 & .71 & \buck{.88}{.57}{.40} & .19 & \buck{.14}{.34}{.34} & 40 & \buck{27}{80}{73} \\
 & \rme & 1 & .75 & \buck{.89}{.61}{.51} & .12 & \buck{.08}{.24}{.19} & 34 & \buck{23}{70}{51} \\
\multirow{-6}{*}{\bf Opus} &  & 2 & .77 & \buck{.92}{.61}{.57} & .11 & \buck{.07}{.21}{.29} & 32 & \buck{17}{76}{79} \\
\hline

 & \ctrlmcp & 1 & .56 & \buck{.72}{.42}{.29} & .06 & \buck{.04}{.08}{.10} & 84 & \buck{68}{114}{133} \\
 &  & 2 & .53 & \buck{.69}{.38}{.26} & .05 & \buck{.05}{.07}{.10} & 84 & \buck{73}{99}{137} \\
 & \rocqmcp & 1 & .82 & \buck{.95}{.67}{.66} & .03 & \buck{.02}{.05}{.05} & 43 & \buck{32}{68}{67} \\
 &  & 2 & .77 & \buck{.90}{.65}{.60} & .03 & \buck{.02}{.06}{.05} & 40 & \buck{25}{80}{59} \\
 & \rme & 1 & .88 & \buck{.97}{.81}{.71} & .02 & \buck{.01}{.03}{.04} & 38 & \buck{26}{62}{71} \\
\multirow{-6}{*}{\bf Terra} &  & 2 & .88 & \buck{.97}{.80}{.71} & .02 & \buck{.01}{.03}{.03} & 30 & \buck{21}{53}{49} \\
\Xhline{1.2pt}

\end{tabular}
\end{small}
\end{center}
\end{table}

\begin{table}[h]
\caption{Results on the five project-scale tasks (four runs per task, two for Opus).}
\label{tab:autoform_results_full}
\begin{center}
\begin{small}
\renewcommand{\arraystretch}{1.2}
\setlength{\tabcolsep}{5pt}
\newcommand{\rot}[1]{\rotatebox[origin=l]{45}{\texttt{#1}}}
\begin{tabular}{llcccccc}
\bf Models & \bf MCP servers & \rot{frugal} & \rot{gauges} & \rot{ledger} & \rot{prodauto} & \rot{triadic} & \rotatebox[origin=l]{45}{\bf Total} \\
\Xhline{1.2pt}

& \ctrlmcp & 4/4 & 3/4 & 3/4 & 0/4 & 0/4 & 10/20 \\
\bf Sonnet & \rocqmcp & 4/4 & 4/4 & 4/4 & 0/4 & 0/4 & 12/20 \\
& \rme & 4/4 & 4/4 & 3/4 & 3/4 & 0/4 & \textbf{14/20} \\
\hline

& \ctrlmcp & 0/2 & 1/2 & 2/2 & 0/2 & 0/2 & 3/10 \\
\bf Opus & \rocqmcp & 2/2 & 2/2 & 2/2 & 0/2 & 0/2 & \textbf{6/10} \\
& \rme & 2/2 & 2/2 & 2/2 & 0/2 & 0/2 & \textbf{6/10} \\
\hline

& \ctrlmcp & 4/4 & 0/4 & 4/4 & 0/4 & 0/4 & 8/20 \\
\bf Terra & \rocqmcp & 3/4 & 2/4 & 2/4 & 0/4 & 0/4 & 7/20 \\
& \rme & 4/4 & 2/4 & 4/4 & 0/4 & 0/4 & \textbf{10/20} \\
\Xhline{1.2pt}

\end{tabular}
\end{small}
\end{center}
\end{table}

\begin{table}[h]
\caption{Results on project-scale tasks with standard deviations: across runs for the accuracy (four runs, two for Opus), over the solved runs for the cost and wall time.}
\label{tab:autoform_results_sd}
\begin{center}
\begin{small}
\renewcommand{\arraystretch}{1.2}
\begin{tabular}{llrrr}
\bf Models & \bf MCP servers & \bf Accuracy & \bf Cost (\$) & \bf Wall time (s) \\
\Xhline{1.2pt}

 & \ctrlmcp & .50 $\pm$ .12 & 1.56 $\pm$ 0.25 & 608 $\pm$ 74 \\
\bf Sonnet & \rocqmcp & .60 $\pm$ .00 & 2.63 $\pm$ 1.05 & 595 $\pm$ 181 \\
 & \rme & .70 $\pm$ .12 & 1.44 $\pm$ 0.84 & 429 $\pm$ 202 \\
\hline

 & \ctrlmcp & .30 $\pm$ .14 & 2.19 $\pm$ 0.12 & 714 $\pm$ 65 \\
\bf Opus & \rocqmcp & .60 $\pm$ .00 & 2.48 $\pm$ 0.27 & 661 $\pm$ 142 \\
 & \rme & .60 $\pm$ .00 & 1.89 $\pm$ 0.74 & 574 $\pm$ 209 \\
\hline

 & \ctrlmcp & .40 $\pm$ .00 & 0.22 $\pm$ 0.12 & 332 $\pm$ 166 \\
\bf Terra & \rocqmcp & .35 $\pm$ .10 & 0.34 $\pm$ 0.19 & 502 $\pm$ 245 \\
 & \rme & .50 $\pm$ .12 & 0.16 $\pm$ 0.11 & 269 $\pm$ 137 \\
\Xhline{1.2pt}

\end{tabular}
\end{small}
\end{center}
\end{table}

\begin{table}[t]
    \caption{Approaches from the PutnamBench leaderboard used to rank the difficulty of the 60 selected problems. The number of stars $\star$ indicates the level of performance of the approaches (based on the number of PutnamBench problems each solves).}
    \label{tab:putnam_selected_models}

    \centering
    \renewcommand{\arraystretch}{1.2}
    \begin{small}
    \begin{tabular}{@{}lll@{}}
      \multicolumn{3}{@{}c@{}}{\bf $\star$} \\
      \Xhline{1.2pt}
      GPT-4o                     & o4-mini-high                & Goedel-Prover-V2 \\
      COPRA (GPT-4o)             & DeepSeek-Prover-V2          & Ax-Prover (Axiomatic AI) \\
      Deepseek R1                & DSP+                        & GPT-5 (ReAct, 10 turns) \\
      Goedel-Prover-SFT          & Bourbaki                    & TIR Conjecturor \\
      ABEL                       & gemini-2.0-flash-thinking    & Enumerate-Conjecture-Prove \\
      InternLM2.5-StepProver     & gemini-2.5-pro-exp-0325   & Kimina-Prover-7B-Distill \\
      Self-play Theorem Prover   &      & \\
      \multicolumn{3}{@{}c@{}}{\bf $\star\star$} \\
      \Xhline{1.2pt}
      AxProverBase (Axiomatic AI)    & Aleph Prover (Logical Intelligence) & Seed-Prover (ByteDance) \\
      Hilbert                    &  & \\
      \multicolumn{3}{@{}c@{}}{\bf $\star\star\star$} \\
      \Xhline{1.2pt}
      Goedel-Architect     & Aleph Prover \#2 (Logical Intelligence) & Seed-Prover 1.5 (ByteDance) \\
      Goedel-Architect \#2 & Aleph Prover \#3 (Logical Intelligence) & \\
    \end{tabular}
    \end{small}
  \end{table}

\begin{table}[h]
\caption{Results of Sonnet on the 60 problems selected from PutnamBench in Lean for each of the two runs. Same conventions as Table~\ref{tab:lean_results}: cost and wall time are averaged over the problems solved by all three MCP servers.}
\label{tab:lean_results_runs}
\begin{center}
\begin{small}
\renewcommand{\arraystretch}{1.2}
\begin{tabular}{l c r@{\hspace{0.45em}}l r@{\hspace{0.45em}}l r@{\hspace{0.45em}}l}
\bf MCP servers & \bf Run & \multicolumn{2}{c}{\bf Accuracy} & \multicolumn{2}{c}{\bf Cost (\$)} & \multicolumn{2}{c}{\bf Wall time (s)} \\
\Xhline{1.2pt}

\ctrlmcp & 1 & .32 & \buck{.90}{.05}{.00} & .12 & \buck{.10}{.31}{--} & 80 & \buck{73}{197}{--} \\
 & 2 & .33 & \buck{.95}{.05}{.00} & .11 & \buck{.11}{--}{--} & 75 & \buck{75}{--}{--} \\
\hline
\leanmcp & 1 & .43 & \buck{.95}{.35}{.00} & .14 & \buck{.11}{.60}{--} & 88 & \buck{75}{298}{--} \\
 & 2 & .40 & \buck{.95}{.25}{.00} & .11 & \buck{.11}{--}{--} & 68 & \buck{68}{--}{--} \\
\hline
\lme & 1 & .35 & \buck{.85}{.20}{.00} & .11 & \buck{.10}{.33}{--} & 62 & \buck{57}{151}{--} \\
 & 2 & .33 & \buck{.80}{.20}{.00} & .08 & \buck{.08}{--}{--} & 51 & \buck{51}{--}{--} \\
\Xhline{1.2pt}

\end{tabular}
\end{small}
\end{center}
\end{table}

\end{document}